\documentclass[10pt,twocolumn,letterpaper]{article}

\usepackage{cvpr}
\usepackage{times}
\usepackage{epsfig}
\usepackage{epstopdf}
\usepackage{graphicx}
\usepackage{amsmath}
\usepackage{amssymb}
\usepackage{helvet}  
\usepackage{courier}  
\usepackage{url}  
\usepackage{graphicx}  
\usepackage{amssymb}
\usepackage{amsmath}
\usepackage{bm}
\usepackage{mathrsfs}
\usepackage{multirow}
\usepackage{enumitem}
\usepackage{listings}

\usepackage[ruled,vlined]{algorithm2e}

\graphicspath{{../picture/}}

\cvprfinalcopy 

\def\cvprPaperID{4648} 

\begin{document}

\title{Aggregation Cross-Entropy for Sequence Recognition}

\author{Zecheng Xie\thanks{Zecheng Xie and Yaoxiong Huang make equal contribution.} , Yaoxiong Huang$^*$, Yuanzhi Zhu, Lianwen Jin\thanks{Corresponding auther.}, Yuliang Liu, Lele Xie\\
South China University of Technology\\
{\tt\small \{zcheng.xie,hwang.yaoxiong,lianwen.jin,zzz.yuanzhi,shaxiaoai18,arlog.lele\}@gmail.com}
}

\maketitle
\thispagestyle{empty}

\begin{abstract}
In this paper, we propose a novel method, aggregation cross-entropy (ACE), for sequence recognition from a brand new perspective.
The ACE loss function exhibits competitive performance to CTC and the attention mechanism, with much quicker implementation (as it involves only four fundamental formulas), faster inference$\backslash$back-propagation (approximately $O(1)$ in parallel), less storage requirement (no parameter and negligible runtime memory), and convenient employment (by replacing CTC with ACE). 
Furthermore, the proposed ACE loss function exhibits two noteworthy properties: (1) it can be directly applied for 2D prediction by flattening the 2D prediction into 1D prediction as the input and (2) it requires only characters and their numbers in the sequence annotation for supervision, which allows it to advance beyond sequence recognition, e.g., counting problem. 
The code is publicly available at \href{https://github.com/summerlvsong/Aggregation-Cross-Entropy}{https://github.com/summerlvsong/Aggregation-Cross-Entropy}.
\end{abstract}


\section{Introduction}
\begin{figure}[t]
\centering
\includegraphics[width=0.48\textwidth]{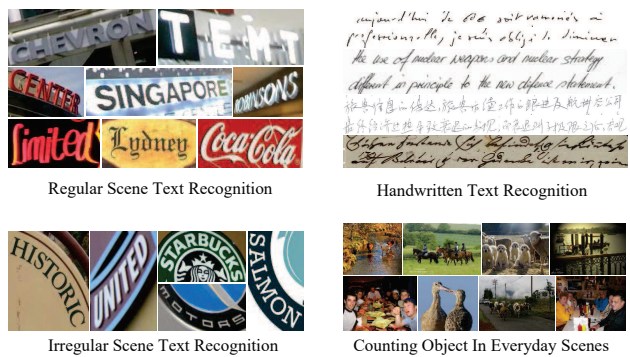}
\caption{Examples of sequence recognition and counting problems.
}
\label{Fig:example}
\end{figure}

Sequence recognition, or sequence labelling \cite{graves2006connectionist} is to assign sequences of labels, drawn from a fixed alphabet, to sequences of input data, e.g., speech recognition\cite{graves2014towards,bahdanau2016end}, scene text recognition \cite{shi2016end,shi2018aster}, and handwritten text recognition \cite{messina2015segmentation,wu2017improving}, as shown in Fig.~\ref{Fig:example}.
The recent advances in deep learning \cite{long2015fully,szegedy2015going,hochreiter1997long} and the new architectures \cite{vaswani2017attention,bluche2016scan,bluche2016joint,wigington2018start} enabled the construction of systems that can handle one-dimensional (1D) \cite{shi2016end,messina2015segmentation} and two-dimensional (2D) prediction problems \cite{zhang2018track,bluche2016joint}.
For 1D prediction problems, the topmost feature maps of the network are collapsed across the vertical dimension to generate 1D prediction \cite{bluche2016scan} because characters in the original images are generally distributed sequentially.
Typical examples are regular scene text recognition \cite{shi2016end,yin2017scene}, online/offline handwritten text recognition \cite{graves2012supervised,messina2015segmentation,wu2017improving}, and speech recognition \cite{graves2014towards,bahdanau2016end}.
For 2D prediction problems, characters in the input image are distributed in a specific spatial structure.
For example, there are highly complicated spatial relations between adjacent characters in mathematical expression recognition \cite{zhang2018track,zhang2017watch}.
In paragraph-level text recognition, characters are generally distributed line by line \cite{bluche2016joint,wigington2018start}, 
whereas in irregular scene text recognition, they are generally distributed in a side-view or curved angle pattern \cite{yang2017learning,cheng2017arbitrarily}.

For the sequence recognition problem, traditional methods generally require to separate training targets for each segment or time-step in the input sequence, resulting in inconvenient pre-segmentation and post-processing stages \cite{graves2012supervised}.
The recent emergence of CTC \cite{graves2006connectionist} and attention mechanism \cite{bahdanau2014neural} significantly alleviate this sequential training problem by circumventing the prior alignment between input image and their corresponding label sequence.
However, although CTC-based networks have exhibited remarkable performance in 1D prediction problem, the underlying methodology is sophisticated; moreover, its implementation, the forward-backward algorithm \cite{graves2012supervised}, is complicated, resulting in large computation consumption. 
Besides, CTC can hardly be applied to 2D prediction problems.
Meanwhile, the attention mechanism relies on its attention module for label alignment, resulting in additional storage requirement and computation consumption.
As pointed out by Bahdanau \emph{et al.} \cite{bahdanau2016end}, recognition model is difficult to learn from scratch with attention mechanism, due to the misalignment between ground truth strings and attention predictions, especially on longer input sequences \cite{kim2017joint,chorowski2015attention}.
Bai \emph{et al.} \cite{bai2018edit} also argues that the misalignment problem can confuse and mislead the training process, and consequently make the training costly and degrade recognition accuracy.
Although the attention mechanism can be adapted for 2D prediction problem, it turns out to be prohibitive in terms of memory and time consumption, as indicated in \cite{bluche2016joint} and \cite{wigington2018start}.

Compelled by the above observations, we propose a novel aggregation cross-entropy (ACE) loss function for the sequence recognition problem, as detailed in Fig.~\ref{Fig:framework}.
Given the prediction of the network, the ACE loss consists of three simple stages: 
(1) aggregation of the probabilities for each category along the time dimension;
(2) normalization of the accumulative result and label annotation as probability distributions over all the classes; and
(3) comparison between these two probability distributions using cross-entropy.
The advantages of the proposed ACE loss function can be summarized as follows:
\begin{itemize}[leftmargin=*, topsep=0pt,itemsep=0pt,parsep=0pt,partopsep=0pt]
\item Owing to its simplicity, the ACE loss function is much quicker to implement (four fundamental formulas), faster to infer and back-propagate (approximately $O(1)$ in parallel), less memory demanding (no parameter and basic runtime memory), and convenient to use (simply replace CTC with ACE), as compared to CTC and attention mechanism. This is illustrated in Table~\ref{tab::investigation}, Section~\ref{complexity_analysis}, and Section~\ref{complexity_result}.
\item Despite its simplicity, the ACE loss function achieves competitive performance to CTC and the attention mechanism, as established in experiments of regular$\backslash$irregular scene text recognition and handwritten text recognition problems.
\item The ACE loss function can be adapted to the 2D prediction problem by flattening the 2D prediction into 1D prediction, as verified in the experiments of irregular scene text recognition and counting problems.
\item The ACE loss function does not require instance order information for supervision, which enable it to advance beyond sequence recognition, e.g., counting problem.
\end{itemize}

\section{Related Work}
\subsection{Connectionist temporal classification}

The advantages of the popular CTC loss were first demonstrated in speech recognition \cite{graves2013speech,graves2014towards} and online handwritten text recognition \cite{graves2009novel,graves2012supervised}.
Recently, an integrated CNN-LSTM-CTC model was proposed to address the scene text recognition problem \cite{shi2016end}.
There are also methods that aim to extend CTC in applications; e.g., Zhang \emph{et al.} \cite{zhang2016application} proposed an extended CTC (ECTC) objective function adapted from CTC to allow RNN-based phoneme recognizers to be trained even when only word-level annotation is available.
Hwang \emph{et al.} \cite{hwang2016sequence} developed an expectation-maximization-based online CTC algorithm that allows RNNs to be trained with an infinitely long input sequence, without pre-segmentation or external reset. 
However, the calculation process of CTC is highly complicated and time-consuming, and it require substantial effort to rearrange the feature map and annotation when applied to 2D problems \cite{wigington2018start,bluche2016joint}.

\subsection{Attention mechanism}
The attention mechanism was first proposed in machine translation \cite{bahdanau2014neural,vaswani2017attention} to enable a model to automatically search for parts of a source sentence for prediction.
Then, the method rapidly became popular in applications such as (visual) question answering \cite{lu2016hierarchical,yang2016stacked},
image caption generation \cite{xu2015show,yang2016stacked,lu2017knowing}, speech recognition \cite{bahdanau2016end,kim2017joint,lu2016hierarchical} and scene text recognition \cite{shi2018aster,bai2018edit,he2016reading}.
Most importantly, the attention mechanism can also be applied to 2D predictions, such as mathematical expression recognition \cite{zhang2018track,zhang2017watch} and paragraph recognition \cite{bluche2016joint,bluche2016scan,wigington2018start}.
However, the attention mechanism relies on a complex attention module to fulfill its functionality, resulting in additional network parameters and runtime. 
Besides, missing or superfluous characters can easily cause misalignment problem, confusing and misleading the training process, and consequently degrading the recognition accuracy \cite{bai2018edit,bahdanau2016end,chorowski2015attention}.

\begin{figure*}[t]
\centering
\includegraphics[width=0.98\textwidth]{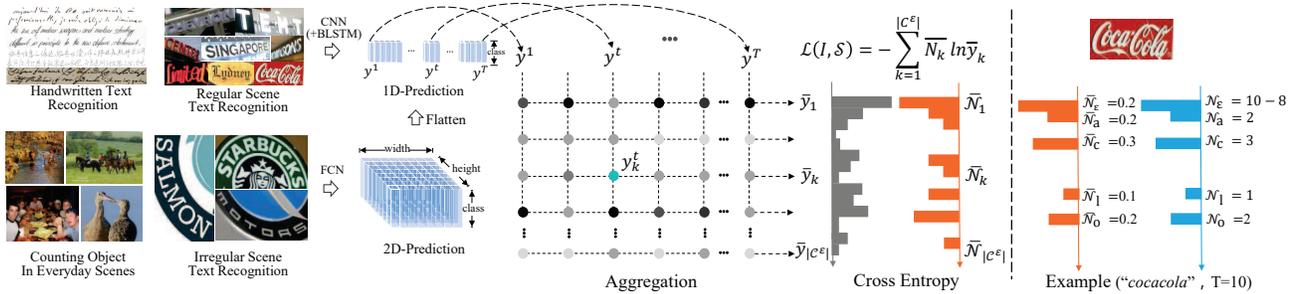}
\caption{(Left) Illustration of proposed ACE loss function.
Generally, the 1D and 2D predictions are generated by integrated CNN-LSTM and FCN model, respectively. 
For the ACE loss function, the 2D prediction is further flattened to 1D prediction, $\{y_k^t, t=1,2,\dots,T\}$.
During aggregation, the 1D predictions at all time-steps are accumulated for each class independently, according to $y^{}_k = \sum^T_{t=1} y_k^t$.
After normalization, the prediction $\overline{y}$, together with the ground-truth $\overline{\mathcal{N}}$, is utilized for loss estimation based on cross-entropy.
(Right) A simple example indicates the generation of annotation for the ACE loss function. $\mathcal{N}_a=2$ implies that there are two ``a'' in \emph{cocacola}.
}
\label{Fig:framework}
\end{figure*}

\section{Aggregation Cross-Entropy}

Formally, given the input image $\mathcal{I}$ and its sequence annotation $\mathcal{S}$ from a training set $\mathcal{Q}$, the general loss function for the sequence recognition problem evaluates the probability of annotation $\mathcal{S}$ of length $L$ conditioned on image $\mathcal{I}$ under model parameter $\omega$ as follows:
\begin{align}
\label{equ:general_loss}
\mathcal{L}(\omega)&=-\sum_{(\mathcal{I}, \mathcal{S}) \in \mathcal{Q}} \log P(\mathcal{S}|\mathcal{I};\omega) \nonumber \\
               &=-\sum_{(\mathcal{I}, \mathcal{S}) \in \mathcal{Q}} \sum_{l=1}^L \log P(\mathcal{S}_l|l,\mathcal{I};\omega)
\end{align}
where $P(\mathcal{S}_l|l,\mathcal{I};\omega)$ represents the probability of predicting character $S_l$ at the $l$-th position of the predicted sequence.
Therefore, the problem is to estimate the general loss function Eq.~\eqref{equ:general_loss} based on the model prediction \{$y^t_k, t = 1, 2, \cdots, T, k = 1, 2, \cdots, |\mathcal{C}^\epsilon|$\}, where $\mathcal{C}^\epsilon = \mathcal{C} \cup \epsilon$, with $\mathcal{C}$ being the character set and $\epsilon$ the blank label.
Nevertheless, directly estimating the probability $P(\mathcal{S}|\mathcal{I};\omega)$ was excessively challenging until the emergence of the popular CTC loss function.
The CTC loss function elegantly calculates $P(\mathcal{S}|\mathcal{I};\omega)$ using a forward-backward algorithm, which removes the need for pre-segmented data and external post-processing.
The attention mechanism provides an alternative solution to estimate the general loss function by directly predicting $P(\mathcal{S}_l|l,\mathcal{I};\omega)$ based on its attention module.
However, the forward-backward algorithm of CTC is highly complicated and time-consuming whereas the attention mechanism requires extra complex network to ensure the alignment between attention prediction and annotation.

In this paper, we present the ACE loss function to estimate the general loss function based on model prediction $y^t_k$.
In Eq.~\eqref{equ:general_loss}, the general loss function can be minimized by maximizing the predictions at each position of the sequence annotation, i.e., $P(\mathcal{S}_l|l,\mathcal{I};\omega)$.
However, directly calculating $P(\mathcal{S}_l|l,\mathcal{I};\omega)$ based on $y^t_k$ is challenging because the alignment between the $l$-th character in the annotation and model prediction $y^t_k$ is unclear.
Therefore, rather than precisely estimating the probability $P(\mathcal{S}_l|l,\mathcal{I};\omega)$, the problem is mitigated by supervising only the accumulative probability of each class; without considering its sequential order in the annotation. 
For example, if a class appears twice in the annotation, we require its accumulative prediction probability over $T$ time-steps to be exactly two, anticipating that its two corresponding predictions approximate to one.
Therefore, we can minimize the general loss function by requiring the network to precisely predict the character number of each class in the annotation as follows:
\begin{align}
\label{equ:ace_s}
\mathcal{L}(\omega)&=-\sum_{(\mathcal{I}, \mathcal{S}) \in \mathcal{Q}} \sum_{l=1}^L \log P(\mathcal{S}_l|l,\mathcal{I};\omega)\nonumber \\
               &\approx -\sum_{(\mathcal{I}, \mathcal{S}) \in \mathcal{Q}} \sum_{k=1}^{|\mathcal{C}^\epsilon|} \log P(\mathcal{N}_{k}|\mathcal{I};\omega)
\end{align}
where $\mathcal{N}_k$ represents the number of times that character $\mathcal{C}^\epsilon_k$ occurs in the sequence annotation $\mathcal{S}$. 
Note that this new loss function does not require character order information but only the classes and their number for supervision.
\subsection{Regression-Based ACE Loss Function}
\label{reg_perspective}

Now, the problem is to bridge model prediction $y^t_k$ to the number prediction of each class.
We propose to calculate the number of each class $y^{}_k$ by summing up the probabilities of the $k$-th characters for $T$ time-steps, i.e., $y^{}_k = \sum^T_{t=1} y_k^t$, as illustrated by \emph{aggregation} in Fig.~\ref{Fig:framework}.
Note that,
\begin{equation}
\max \sum_{k=1}^{|\mathcal{C}^\epsilon|} \log P(\mathcal{N}_{k}|\mathcal{I};\omega) \Leftrightarrow \min \sum^{|\mathcal{C}^\epsilon|}_{k=1}(\mathcal{N}_k -y_k)^2
\end{equation}
Therefore, we adapt the loss function (Eq.~\eqref{equ:ace_s}) from the perspective of regression problem as follows:
\begin{align}
\label{equ:regression}
\mathcal{L}(\omega)=\frac{1}{2}\sum_{(\mathcal{I}, \mathcal{S}) \in \mathcal{Q}} \sum^{|\mathcal{C}^\epsilon|}_{k=1}(\mathcal{N}_k -y_k)^2.                          
\end{align}
Also note that a total of $(T-|\mathcal{S}|)$ predictions are expected to yield null emission. Therefore, we have $\mathcal{N}_\epsilon=T-|\mathcal{S}|$.

To find the gradient for each example $(\mathcal{I},\mathcal{S})$, we first differentiate $\mathcal{L}(\mathcal{I},\mathcal{S})$ with respect to the network output $y_k^t$:
\begin{align}
\label{equ:gradient}
\frac{\partial \mathcal{L}(\mathcal{I},\mathcal{S})}{\partial y_k^t}&= \Delta_k,
\end{align}
where $\Delta_k = (y_k -\mathcal{N}_k)$.
Recall that for Softmax functions, we have:
\begin{align}
\label{equ:softmax}
y_i=\frac{e^{a_i}}{\sum_{j}e^{a_{j}}},\ 
\frac{\partial y_i}{\partial a_j} = y_{i}(\delta_{ij}- y_j),
\end{align}
where $\delta_{ij} = 1$ if $i=j$ and zero otherwise.
Now, we can differentiate the loss function with respect to $a_k^t$ to back-propagate the gradient through the output layer:
\begin{align}
\label{equ:regression_derivative}
\frac{\partial \mathcal{L}(\mathcal{I},\mathcal{S})}{\partial a_k^t} &= \sum_{k'=1}^{|\mathcal{C}^\epsilon|} \frac{\partial \mathcal{L}(\mathcal{I},\mathcal{S})}{\partial y_{k'}^t}\frac{\partial y_{k'}^t}{\partial a^{t}_k} \nonumber 
                                                    = \sum_{k'=1}^{|\mathcal{C}^\epsilon|} \Delta_{k'} \cdot y_{k'}^t(\delta_{kk'}- y_k^t) \nonumber \\
                                                    &= \Delta_{k'} \cdot y_{k'}^t(1- y_k^t) - \sum_{k'\neq k} \Delta_{k'} \cdot y_{k'}^t y_k^t 
\end{align}
\subsubsection{Gradient vanishing problem}
From Eq.~\eqref{equ:regression_derivative}, we observe that the regression-based ACE loss (Eq.~\eqref{equ:regression}) is not convenient in term of back-propagation.
In the early training stage, we have $\{y_{k'}^t\approx 1/|\mathcal{C^\epsilon}|, \forall k', t\}$.
Therefore, $y_{k'}^t$ will be negligible for large vocabulary sequence recognition problems, where $|\mathcal{C^\epsilon}|$ is large (e.g., 7,357 for the HCTR problem). 
Although the other terms in Eq.~\eqref{equ:regression_derivative} (e.g., $\Delta_{k'}$) have acceptable magnitudes for back-propagation, the gradient would be scaled to a remarkably small size by the term $y_{k'}^t$ and $y_{k}^t$, resulting in gradient vanishing problem.



\subsection{Cross-Entropy-Based ACE Loss Function}
\label{ce_perspective}
To prevent the gradient-vanishing problem, It is necessary to offset the negative effect of the term $y_{k'}^t$ introduced by the Softmax function in Eq.~\eqref{equ:regression_derivative}.
We borrow the concept of \emph{cross-entropy} from information theory, which is designed to measure the ``distance'' between two probability distributions. 
Therefore, we first normalize the accumulative probability of the $k$-th character $y_k$ to $\overline{y}_k=y_k/T$, and the character numbers $\mathcal{N}_k$ to $\overline{\mathcal{N}}_k=\mathcal{N}_k/T$.
Then, the cross-entropy between $\overline{y}$ and $\overline{\mathcal{N}}$ is expressed as:
\begin{align}
\label{equ:cross_entropy}
\mathcal{L}(\mathcal{I},\mathcal{S}) =-\sum^{|\mathcal{C}^\epsilon|}_{k=1}\overline{\mathcal{N}}_{k}\ln \overline{y}_k
\end{align}

The loss function derivatives with respect to $a_k^t$ before the Softmax activation function has the following form:
\begin{align}
\label{equ:ce_gradient}
\frac{\partial \mathcal{L}(\mathcal{I},\mathcal{S})}{\partial a_k^t} &= \sum_{k'=1}^{|\mathcal{C^\epsilon}|}\frac{\partial \mathcal{L}(\mathcal{I},\mathcal{S})}{\partial \overline{y}_{k'}}\frac{\partial \overline{y}_{k'}}{\partial y_{k'}^t}\frac{\partial {y}_{k'}^t}{\partial a_k^t}\nonumber \\
                                                    &= \sum_{\mathcal{C}^\epsilon_{k'} \in \mathcal{S}}-\frac{\overline{\mathcal{N}}_{k'}}{\overline{y}_{k'}}\cdot \frac{1}{T}\cdot y_{k'}^t(\delta_{kk'}- y_k^t) \nonumber \\
                                                    &= -\frac{1}{T} \sum_{\mathcal{C}^\epsilon_{k'} \in \mathcal{S}}\overline{\mathcal{N}}_{k'}\frac{y_{k'}^t}{\overline{y}_{k'}} (\delta_{kk'}- y_k^t)\nonumber \\
                                                    &= -\frac{1}{T} \sum_{\mathcal{C}^\epsilon_{k'} \in \mathcal{S}}\overline{\mathcal{N}}_{k'}\frac{y_{k'}^t}{\overline{y}_{k'}} (\delta_{kk'}- y_k^t)
\end{align}
\subsubsection{Discussion}
In the following, we explain how the updated loss function solves the gradient vanishing problem:

(1) In the early training stage, $y_{k'}^t$ has an approximately identical order of magnitude at all the time-steps. Thus, the normalized accumulated probability $\overline{y}_{k'}$ is also of an identical order of magnitude as $y_{k'}^t$.
That is, $\frac{y_{k'}^t}{\overline{y}_{k'}}\approx 1$; therefore, the gradient through the $k'$-th class is now $-\frac{1}{T}\overline{\mathcal{N}}_{k'}(\delta_{kk'}- y_k^t)$.
Thus, the gradient can straightforwardly back-propagate to $a_k^t$ through the characters that appear in sequence annotation $\mathcal{S}$.
Besides, when $k=k'$, i.e., $\mathcal{C}^\epsilon_k \in \mathcal{S}$; the corresponding gradient is approximately  $-\frac{1}{T}\overline{\mathcal{N}}_{k'}(1- y_k^t)$, which will encourage the model to make a larger prediction $y_k^t$, whereas characters that do not appear in $\mathcal{S}$ become smaller.
This was our original intention.

(2) In the later training stage, only a few of the prediction $y^{t^*}_{k^*}$ will be very large, leaving the other predictions small enough to be omitted.
In this situation, prediction $y^{t^*}_{k^*}$ will occupy the majority of $y^{}_{k^*}$, and we have $\frac{y_{k^*}^{t^*}}{\overline{y}_{k^*}}=T \cdot \frac{y_{k^*}^{t^*}}{y_{k^*}}$.
Therefore, when $\mathcal{C}^\epsilon_{k*} \in \mathcal{S}$, the gradient can be straightforwardly back-propagated to the recognition network.





\subsection{Two-dimensional Prediction}
In some 2D prediction problem like irregular scene text recognition with image level annotations, it is challenging to define the spatial relation between characters.
Characters may be arranged in multiple lines, in a curved or sloped direction, or even distributed in a random manner.
Fortunately, the proposed ACE loss function can naturally be generalized for the 2D prediction problem, because it does not require character-order information for the sequence-learning process.

Suppose that the output 2D prediction ${\bm y}$ has height $\mathcal{H}$ and width $\mathcal{W}$, and the prediction at the $h$-th line and $w$-th row is denoted as $y_k^{hw}$.
This requires a marginal adaptation of the calculation of $\overline{y}_k$ and $\overline{\mathcal{N}}_k$ as follows, $\overline{y}_k = \frac{y_k}{\mathcal{H}\mathcal{W}} = \frac{\sum_{h=1}^{\mathcal{H}}\sum_{w=1}^{\mathcal{W}}y_k^{hw}}{\mathcal{H}\mathcal{W}}$, $\overline{\mathcal{N}}_k = \frac{\mathcal{N}_k}{\mathcal{H}\mathcal{W}}$.
Then, the loss function for the 2D prediction can be transformed as follows:
\begin{align}
\label{equ:cs_twodim}
\mathcal{L}(\mathcal{I},\mathcal{S}) &=-\sum_{k=1}^{|\mathcal{C^\epsilon}|}\overline{\mathcal{N}}_{k}\ln \overline{y}_k =-\sum_{k=1}^{|\mathcal{C^\epsilon}|}\frac{\mathcal{N}_k}{\mathcal{H}\mathcal{W}}\ln \frac{y_k}{\mathcal{H}\mathcal{W}}
\end{align}
In our implementation, we directly flatten the 2D prediction $\{ y^{hw}, h = 1,2,\cdots,\mathcal{H}, w = 1,2,\cdots,\mathcal{W}\}$ into 1D prediction $\{y^{t}, t = 1,2,\cdots, T\}$, where $T=\mathcal{HW}$, and then apply Eq.~\eqref{equ:cross_entropy} to calculate the final loss.

\subsection{Implementation and Complexity Analysis}
\label{complexity_analysis}
{\bf Implementation}
As illustrated in Eq.~\eqref{equ:ace_s}, $\mathcal{N} = \{\mathcal{N}_k|k=1,2,\cdots,|\mathcal{C}^\epsilon|\}$ represents the annotation for the ACE loss function; here, $\mathcal{N}_k$ represents the number of times that the character $\mathcal{C}^\epsilon_k$ occurs in the sequence annotation $\mathcal{S}$.
A simple example describing the translation of sequence annotation \emph{cocacola} into ACE's annotation is shown in Fig.~\ref{Fig:framework}.
In conclusion, given the model prediction $y_k^t$ and its annotation $\mathcal{N}$, the key implementation for a cross-entropy-based ACE loss function consists of four fundamental formulas: 
\begin{itemize}[leftmargin=*, topsep=0pt,itemsep=1pt,parsep=0pt,partopsep=0pt]
\item $y^{}_k = \sum^T_{t=1} y_k^t$ to calculate the character number of each class by summing up the probabilities of the $k$-th class for all $T$ time-steps.
\item $\overline{y}_k=y_k/T$ to normalize the accumulative probabilities.
\item $\overline{\mathcal{N}}_k=\mathcal{N}_k/T$  to normalize the annotation.
\item $\mathcal{L}(\mathcal{I},\mathcal{S})=-\sum^{|\mathcal{C}^\epsilon|}_{k=1}\overline{\mathcal{N}}_{k}\ln \overline{y}_k$ to estimate the cross-entropy between $\overline{\mathcal{N}}_k$ and $\overline{y}_k$. 
\end{itemize}

In practical employment, the model prediction $y_k^t$ is generally provided by the integrated CNN-LSTM model (1D prediction) or FCN model (flattened 2D prediction).
That is, the input assumption of ACE is identical to that of CTC; therefore, the proposed ACE can be conveniently applied by replacing the CTC layer in the framework.

{\bf Complexity Analysis}
The overall computation of the ACE loss function is implemented based on the above-mentioned four formulas that have computation complexities of $O(1)$, $O(|\mathcal{C}^\epsilon|)$, $O(|\mathcal{C}^\epsilon|)$, and $O(|\mathcal{C}^\epsilon|)$, respectively. Therefore, the computation complexity of the ACE loss function is $O(|\mathcal{C}^\epsilon|)$. 
Note however that the element-wise multiplication, division, and log operation in these four formulas can be implemented in parallel with GPU at $O(1)$.
In contrast, the implementation of CTC \cite{graves2012supervised} based on a forward-backward algorithm has a computation complexity of $O(T*|\mathcal{S}|)$.
Because the \emph{forward variable} $\alpha(t,u)$ and \emph{backward variable} $\beta(t,u)$ \cite{graves2012supervised} of CTC depend on the previous result (e.g., $\alpha(t-1,u)$ and $\beta(t+1,u)$) to calculate the present output, CTC can hardly be accelerated in parallel in the time dimension.
Moreover, the elementary operation $\alpha(t,u)$ of CTC is already very complicated, resulting in larger overall time consumption than that of ACE.
With regard to the attention mechanism, its computation complexity is proportional to the times of `attention'. 
However, the computation complexity of the attention module at each time already has similar magnitude as that of CTC.

From the perspective of memory consumption, the proposed ACE loss function requires nearly no memory consumption because the ACE loss result can be directly calculated based on the four fundamental formulas.
However, CTC requires additional space to preserve the  \emph{forward$\backslash$backward variable} that is proportional to the time-step $T$ and the length of the sequence annotation.
Meanwhile, the attention mechanism requires additional module to implement ``attention''. Thus, its memory consumption is significantly larger than that of CTC and ACE.

In conclusion, the proposed ACE loss function exhibits significant advantages with regard to both computation complexity and memory demand, as compared to CTC and attention.


\section{Performance Evaluation}
\label{sec:per_eval}
In our experiment, three tasks were employed to evaluate the effectiveness of the proposed ACE loss function, including scene text recognition,  offline handwritten Chinese text recognition, and counting objects in everyday scenes.
For these tasks, we estimated the ACE loss for 1D and 2D predictions, where 1D implies that the final prediction is a sequence of T predictions and 2D indicates that the final feature map has 2D predictions of shape $\mathcal{H}\times \mathcal{W}$.

\subsection{Scene Text Recognition}
Scene text recognition often encounter problems owing to the large variations in the background, appearance, resolution, text font, and color, making it a challenging research topic.
In this section, we study both 1D and 2D predictions on scene text recognition by utilizing the richness and variety of the testing benchmark in this task.

\subsubsection{Dataset} Two types of datasets are used for scene text recognition: regular text datasets, such as IIIT5K-Words \cite{mishra2012scene}, Street View Text \cite{wang2011end}, ICDAR 2003 \cite{lucas2005icdar}, and ICDAR 2013 \cite{karatzas2013icdar}, and irregular text datasets, such as SVT-Perspective \cite{quy2013recognizing}, CUTE80 \cite{risnumawan2014robust}, and ICDAR 2015 \cite{karatzas2015icdar}.
The regular datasets were used to study the 1D prediction for the ACE loss function while the irregular text datasets were applied to evaluate the 2D prediction.

IIIT5K-Words (\emph{IIIT5K}) contains 3000 cropped word images for testing. 

Street View Text (\emph{SVT}) was collected from Google Street View, including 647 word images. Many of them are severely corrupted by noise and blur, or have very low resolutions. 

ICDAR 2003 (\emph{IC03}) contains 251 scene images that are labeled with text bounding boxes. The dataset contains 867 cropped images.

ICDAR 2013 (\emph{IC13}) inherits most of its samples from IC03. It contains 1015 cropped text images. 

SVT-Perspective (\emph{SVT-P}) contains 639 cropped images for testing, which are selected from side-view angle snapshots from Google Street View. Therefore, most of images are perspective distorted. Each image is associated with a 50-word lexicon and a full lexicon.

CUTE80 (\emph{CUTE}) contains 80 high-resolution images taken of natural scenes. It was specifically collected for curve text recognition. The dataset contains 288 cropped natural images for testing. No lexicon is associated.

ICDAR 2015 (\emph{IC15}) contains 2077 cropped images including more than 200 irregular text.

\subsubsection{Implementation Details}
For 1D sequence recognition on regular datasets, our experiments were based on the CRNN \cite{shi2016end} network, trained only on 8-million synthetic data released by Jaderberg \emph{et al.} \cite{jaderberg2014synthetic}.
For 2D sequence recognition on irregular datasets, our experiments were based on the ResNet-101 \cite{he2016deep}, with conv1 changed to 3$\times$3, stride 1, and conv4\_x as output.
The training dataset consists of 8-million synthetic data released by Jaderberg \emph{et al.} \cite{jaderberg2014synthetic} and 4-million synthetic instances (excluding the images that contain non-alphanumeric characters) cropped from 80-thousand images \cite{gupta2016synthetic}.
The input images are normalized to the shape of (96,100) and the final 2D prediction has the shape of (12,13), as shown in Fig.~\ref{fig:curve}.
To decode the 2D prediction, we flattened the 2D prediction by concatenating each column in order from left to right and top to bottom and then decoded the flattened 1D prediction following the general procedure.

In our experiment, we observed that directly normalizing the input image to the size of (96,100) overloads the network training process.
Therefore, we trained another network to predict the character number in the text image and normalized the text image with respect to the character number to keep the character size within acceptable limits.

\begin{table}[!t]
\center
\small
\caption{Comparison between regression and cross-entropy.}
\label{tab:comparison_reg_ce}
\begin{tabular}{c|cccc}
\hline                        
Method &IIIT5K& SVT& IC03 & IC13        \\ \hline
Shi \emph{et al.} \cite{shi2016end}                  & 81.2                    & \textbf{82.7}                 & 91.9          & 89.6 \\ 
ACE (1D, Regression)                 & 19.4                   & 6.6           & 12.0 & 9.3\\ 
ACE (1D, Cross Entropy)               & \textbf{82.3}                   & 82.6           & \textbf{92.1} & \textbf{89.7}\\ \hline
\end{tabular}
\end{table}

\subsubsection{Experimental Result}
\begin{figure}[!b]
\includegraphics[width=0.46\textwidth]{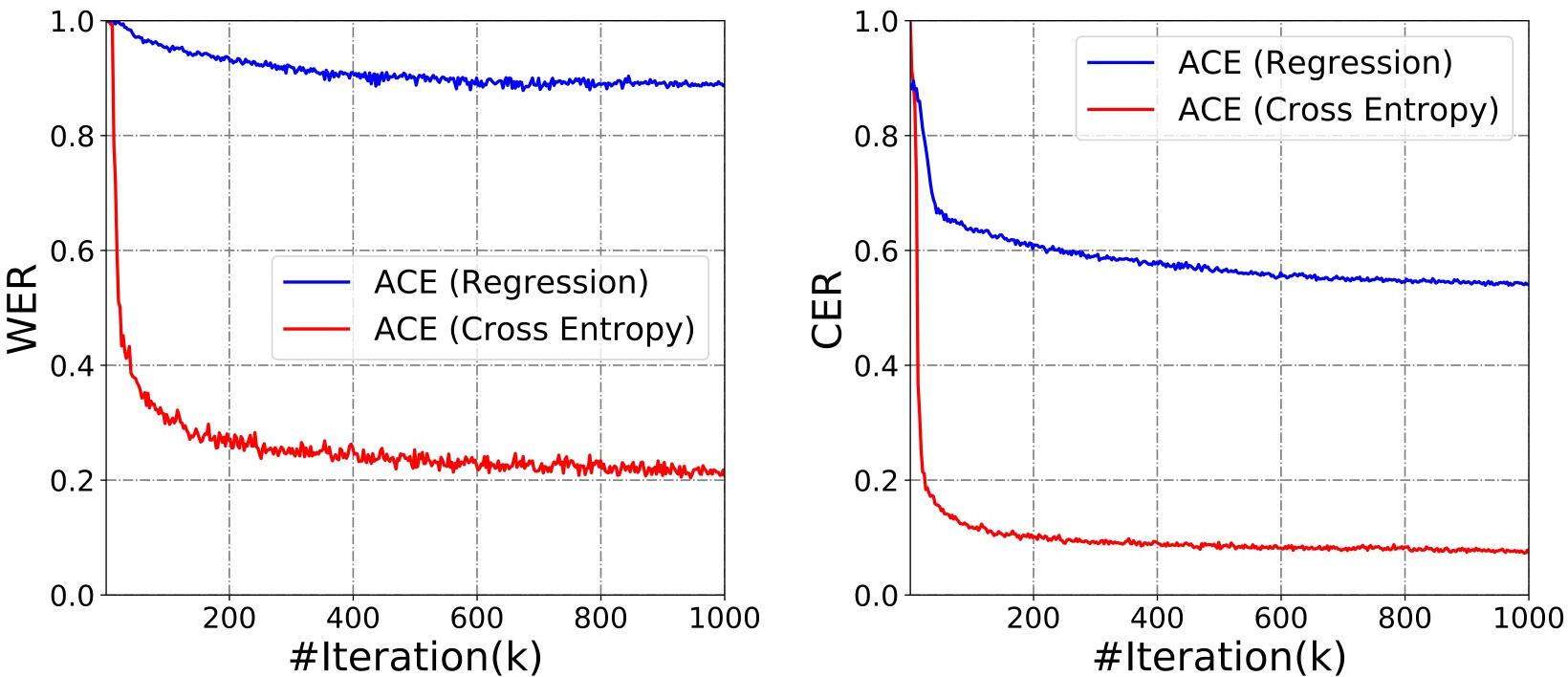}
\label{comparison_reg_ce}
\caption{
   Word error rate (left) and character error rate (right) of ACE loss on validation set under regression and cross entropy perspective.
}
\label{fig:compare_reg_ce}
\end{figure}

To study the role of regression and cross-entropy for the ACE loss function, we conducted experiments with 1D prediction using regular scene text datasets, as detailed in Table \ref{tab:comparison_reg_ce} and Fig.~\ref{fig:compare_reg_ce}.
Because there are only 37 classes in scene text recognition, the negative effect of the term $y^t_{k'}$ in Eq.~\eqref{equ:regression_derivative} is not as significant as that of the HCTR problem (7357 classes).
As shown in Fig.~\ref{fig:compare_reg_ce}, with regression-based ACE loss, the network can converge but at a relatively slow rate, probably due to the gradient vanishing problem.
With cross-entropy-based ACE loss, the WER and CER evolve at a relatively higher rate and in a smoother manner at the early training stage and attain a significantly better convergence result in the subsequent training stage.
Table \ref{tab:comparison_reg_ce} clearly reveals the superiority of the cross-entropy-based ACE loss function over the regression-based one. 
Therefore, we use cross-entropy-based ACE loss functions for all the remaining experiments.
Moreover, with the same network setting (CRNN) and training set (8-million synthetic data), the proposed ACE loss function exhibits performance comparable with that of previous work \cite{shi2016end} with CTC.

\begin{figure}[!b]
\includegraphics[width=0.46\textwidth]{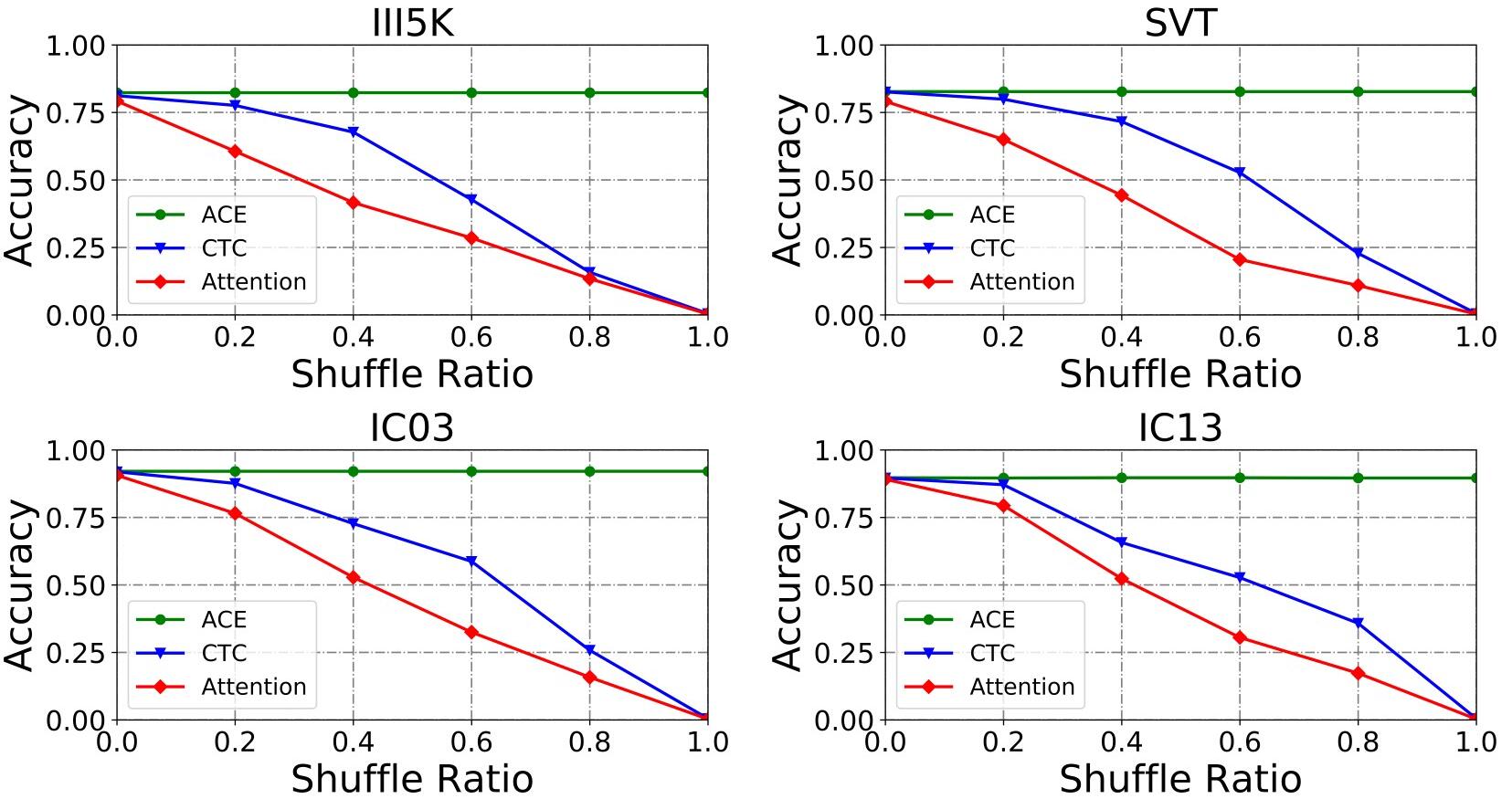}
\caption{Performance of ACE, CTC, and attention under different shuffle ratios and datasets.
}
\label{fig:shuffle}
\end{figure}

To validate the independence of the proposed ACE loss to character order, we conduct experiments with ACE, CTC, and attention on four datasets; the character order of annotation is randomly shuffled at different ratios, as shown in Fig.~\ref{fig:shuffle}.
It is observed that the performance of attention and CTC on all the datasets degrades as the shuffle ratio increases.
Specifically, attention is more sensitive than CTC because misalignment problem can easily misleads the training process of attention \cite{bai2018edit}.
In contrast, the proposed ACE loss function exhibits similar recognition results for all the settings of the shuffle ratio, this is because it only requires classes and their number for supervision, completely omitting character order information.

\begin{table}[!t]
\center
\small
\caption{Comparison with previous methods for scene text recognition problem (without rectification) }
\label{comparison_scene_text}
\begin{tabular}{@{}l|c|ccc|c|c@{}}
\hline                        
\multirow{2}{*}{Method} & \multirow{2}{*}{2D}&           \multicolumn{3}{c|}{SVT-P}           & CUTE        & IC15          \\
\cline{3-7}
                                            && 50            & Full          & None          & None          & None          \\
\hline                        
Shi \emph{et al.} \cite{shi2016end}                           && 92.6          & 72.6          & 66.8          & 54.9  &-  \\
Liu \emph{et al.} \cite{liu2016star}                                  && 94.3          & 83.6          & 73.5          & -      &-                    \\
Yang \emph{et al.} \cite{yang2017learning}             &\checkmark& 93.0          & 80.2          & \textbf{75.8} & 69.3          & -             \\
Cheng \emph{et al.} \cite{cheng2017focusing}                      &\checkmark& 92.6          & 81.6          & 71.5          & 63.9          & 66.2          \\
Cheng \emph{et al.} \cite{cheng2017arbitrarily}            &\checkmark& 94.0          & 83.7          & 73.0          & 76.8          & 68.2          \\
Liu \emph{et al.} \cite{liu2018synthetically}           && --          & --          & 73.9          & 62.5          & --          \\
Shi \emph{et al.} \cite{shi2018aster}           && --          & --          & 74.1          & 73.3          & --          \\
\hline
ACE (2D)    &\checkmark& \textbf{94.9} & \textbf{87.8} & 70.1          & \textbf{82.6} & \textbf{68.9}\\
\hline
\end{tabular}
\end{table}

For irregular scene text recognition, we conducted text recognition experiments with 2D prediction. 
In Table \ref{comparison_scene_text}, we provide a comparison with previous methods that considered only recognition model and no rectification for fair comparison.
As illustrated in Table \ref{comparison_scene_text}, the proposed ACE loss function exhibits superior performance on the datasets CUTE and IC15, particularly on CUTE with an absolute error reduction of 5.8\%.
This is because the dataset CUTE was specifically collected for curved text recognition and therefore, fully demonstrates the advantages of the ACE loss function.
For the dataset SVT-P, our naive decoding result is less effective than that of Yang \emph{et al.} \cite{yang2017learning}.
This is because numerous images in the dataset SVT-P have very low resolutions, which creates a very high requirement for semantic context modeling.
However, our network is based only on CNN, with neither LSTM/MDLSTM nor attention mechanism to leverage the high-level semantic context.
Nevertheless, it is noteworthy that our recognition model achieved the highest result when using lexicons, with which semantic context is accessible.
This again validates the robustness and effectiveness of the proposed ACE loss function.

\begin{figure}[!htb]
\centering
\includegraphics[width=0.45\textwidth]{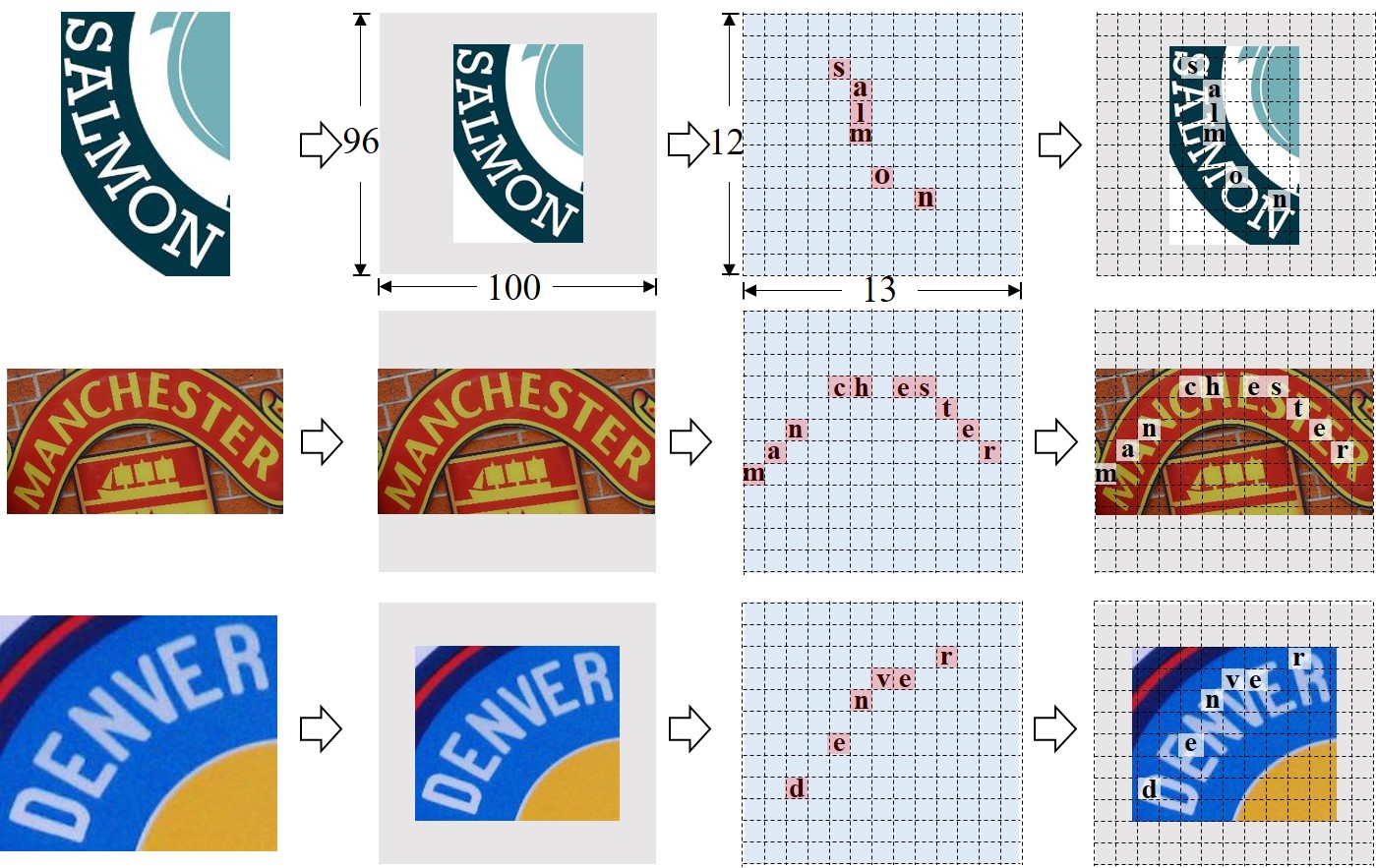}
\caption{Real images processed by recognition model using ACE loss function. The left two columns represent original text images and their normalized versions within the shape of (96, 100).
The third column shows the 2D prediction of the recognition model for the text images.
In the right column, we overlap the input and the prediction images, and observe similar character distribution in the 2D space.
}
\label{fig:curve}
\end{figure}

In Fig.~\ref{fig:curve}, we provide a few real images processed by a recognition model using the ACE loss function. The original text images were first normalized and placed in the center of a blank image of shape (96, 100). 
We observe that after recognition, the 2D prediction exhibits a spatial distribution highly similar to that of the characters in the original text image, which implies the effectiveness of the proposed ACE loss function.

\subsection{Offline Handwritten Chinese Text Recognition}
\label{HCTR}
Owing to its large character set (7,357 classes), diverse writing style, and character-touching problem, the offline HCTR problem is highly complicated and challenging to solve. Therefore, it is a favorable testbed to evaluate the robustness and effectiveness of the ACE loss in 1D predictions.

\subsubsection{Implementation Details}

For the offline HCTR problem, our model was trained using the CASIA-HWDB \cite{liu2011casia} datasets and tested with the standard benchmark ICDAR 2013 competition dataset \cite{yin2013icdar}.





For the HCTR problem, our network architecture with a prediction sequence of length 70 is specified as follows:\\
{\small $(126,576)Input-8C3-MP2-32C3-MP2-128C3-MP2-5*256C3-MP2-512C3-512C3-MP2-512C2-3*512ResLSTM-7357FC-Output$},\\
where $xCy$ represents a convolutional layer with kernel number of $x$ and kernel size of $y*y$, $MPy$ denotes a max pooling layer with kernel size of $y$, and $xFC$ is a fully connected layer with kernel number of $x$, and ResLSTM is residual LSTM proposed in \cite{xie2018learning}.
The evaluation criteria for the HCTR problem are correct rate (CR) and accuracy rate (AR) specified by ICDAR2013 competition \cite{yin2013icdar}.


\subsubsection{Experimental Result}

In Table~\ref{tab::HCTR_comparison}, we provide the comparison between ACE loss and previous methods. 
It is evident that the proposed ACE loss function exhibits higher performance than previous methods, including MDLSTM-based models \cite{messina2015segmentation,wu2017handwritten}, HMM-based model \cite{du2016deep}, and over-segmentation methods \cite{liuicdar,wang2012handwritten,wang2016deep,wu2017improving} with and without language model (LM).
Compared to scene text recognition, handwritten Chinese text recognition problem possesses its unique challenges, such as large character set (7357 classes) and character-touching problem.
Therefore, the superior performance of ACE loss function over previous methods can properly verify its robustness and generality for sequence recognition problems.

\begin{table}[!h]
\center
\small
\caption{Comparison with previous methods for HCTR. }
\label{tab::HCTR_comparison}
\begin{tabular}{l|cc|cc}
\hline
\multirow{2}{*}{Method} & \multicolumn{2}{c|}{w.o LM }       & \multicolumn{2}{c}{with LM }       \\ \cline{2-5}
         & CR & AR & CR & AR\\ \hline
HIT-2 \cite{liuicdar}      & -- & -- & 88.76 & 86.73    \\ 
Wang \emph{et al.} \cite{wang2012handwritten}      & -- & -- & 91.39 & 90.75    \\ 
Messina \emph{et al.} \cite{messina2015segmentation}      & -- & 83.50 & -- & 89.40    \\ 
Wu \emph{et al.} \cite{wu2017handwritten}       & 87.43 & 86.64 & -- & 92.61    \\ 
Du \emph{et al.} \cite{du2016deep}       & -- & 83.89 & -- & 93.50 \\ 
Wang \emph{et al.} \cite{wang2016deep}       & 90.67 & 88.79 & 95.53 & 94.02 \\ 
Wu \emph{et al.} \cite{wu2017improving}       & -- & --       & 96.32 &   96.20 \\ \hline
ACE (1D)    &   {\bf 91.68} & {\bf 91.25}      & {\bf 96.70}           & {\bf 96.22}       \\ \hline
\end{tabular}
\end{table}


\subsection{Counting Objects in Everyday Scenes}
Counting the number of instances of object classes in natural everyday images generally encounters complex real life situations, e.g., large variance in counts, appearance, and scales of object.
Therefore, we verified the ACE loss function on the problem of counting objects in everyday scenes to demonstrate its generality.

\subsubsection{Implementation Details}
As a benchmark for multi-label object classification and object detection tasks, the PASCAL VOC \cite{everingham2015pascal} datasets contain category labels per image, as well as bounding box annotations that can be converted to the object number labels. 
In our implementation, we accumulated the prediction for category $k$ to obtain $\hat{c}_{ik}$ by thresholding counts at zero and rounding predictions to the closest integers.
Given these predictions and the ground truth counts $c_{ik}$ for a category $k$ and image $i$, \emph{RMSE} and \emph{relRMSE} is calculated by $\emph{RMSE}_k= \sqrt{\frac{1}{N}\sum^N_{i=1}(\hat{c}_{ik}-c_{ik})^2}$ and $\emph{relRMSE}_k= \sqrt{\frac{1}{N}\sum^N_{i=1}\frac{(\hat{c}_{ik}-c_{ik})^2}{c_{ik}+1}}$.

\subsubsection{Experimental Result}
Table~\ref{tab:comp_counting} presents a comparison between the proposed ACE loss function and previous methods for the PASCAL VOC 2007 test dataset for counting objects in everyday scenes.
The proposed ACE loss function outperforms the previous glancing and subitizing method \cite{chattopadhyay2017counting}, correlation loss method \cite{song2018learn}, and Always-0 method (predicting most-frequent ground truth count).
The results have shown the generality of ACE loss function, in that it can be readily applied to problem other than sequence recognition, e.g., counting problems, requiring minimal domain knowledge.

\begin{figure}[!h]
\centering
\includegraphics[width=0.5\textwidth]{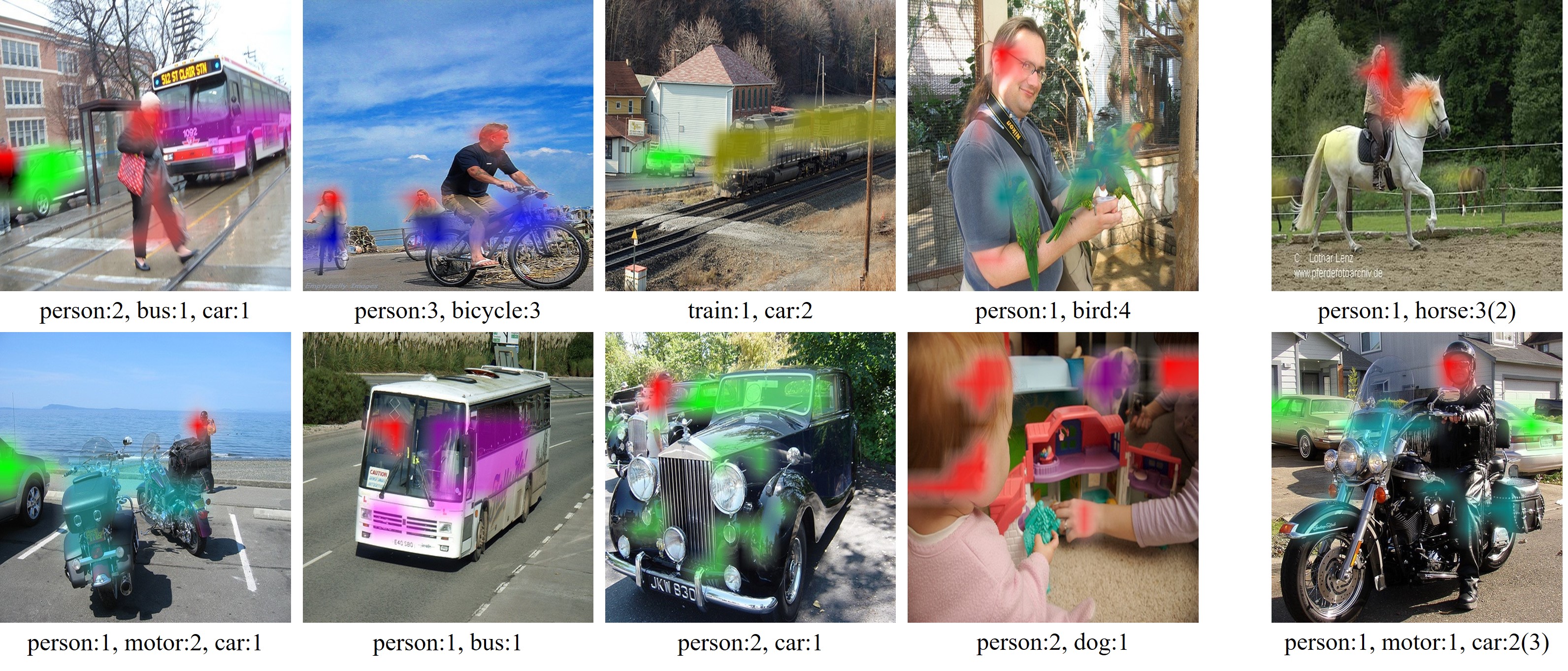}
\caption{
   Real images processed by counting model using ACE loss function. The first four columns display examples that are correctly recognized by our model.
   The top-right image is correctly recognized, but with an incorrectly annotated label. 
   (Incorrect predictions are provided with labels in brackets)
}
\label{fig:counting}
\end{figure}

In Fig.~\ref{fig:counting}, we provide real images processed by the counting model under ACE loss. 
As shown in the images, our counting model trained with ACE loss manage to pay ``attention'' to the position where crucial objects occur. 
Unlike the text recognition problem, where the recognition model trained with the ACE loss function tends to make a prediction for a character, the counting model trained with the ACE loss function provides a more uniform prediction distribution over the body of the object.
Moreover, it assigns different levels of ``attention'' to different parts of an object. 
For example, when observing the red color in the pictures, we notice that the counting model pays more attention to the face of a person.
This phenomenon corresponds to our intuition because the face is the most distinctive part of an individual.

\begin{table}[!h]
\small
\center
\caption{Comparison with previous methods on PASCAL VOC 2007 test dataset for object counting problem.}
\label{tab:comp_counting}
\begin{tabular}{l|cc}
\hline
Method                                                                           & \small{m-RMSE} & \small{m-relRMSE} \\ \hline
Always-0                                                                         & 0.665 & 0.284     \\ 
Glance \cite{chattopadhyay2017counting}                                                                        & 0.500 & 0.270     \\ 
Sub-ens  \cite{chattopadhyay2017counting}                                                                        & 0.420 & 0.200     \\ 
Two-stream \cite{song2018learn} &  0.389 &  0.189     \\ \hline
ACE (2D) & {\bf 0.381} & {\bf 0.185}     \\ \hline
\end{tabular}
\end{table}

\subsection{Complexity Analysis}
\label{complexity_result}
In Table ~\ref{tab::investigation}, we compare the parameter, runtime memory, and run time of ACE with those of CTC and attention.
The result is executed with minibatch 64 and model prediction length T=144 on a single NVIDIA TITAN X graphics card of 12GB memory.
Similar to CTC, the proposed ACE does not require any parameter to fulfill its function.
Owing to its simplicity, ACE requires marginal runtime memory, five times less than those for CTC and attention.
Furthermore, its speed is as least 30 times higher than those of CTC and attention.
It is note worthy that with all these advantages, the proposed ACE achieve performance that is comparable or higher than those with CTC and attention.

\begin{table}[!h]
\small
\center
\caption{Investigation over parameter (Para), runtime memory (Mem), and speed (Speed) (in units of MB, MB, and ms, respectively) of CTC, attention, and ACE.}
\label{tab::investigation}
\begin{tabular}{c|ccc|ccc}
\hline
\multirow{2}{*}{Method}           & \multicolumn{3}{c|}{37 classes}       & \multicolumn{3}{c}{7357 classes}       \\ 
\cline{2-7}
                                                             & Para & Mem  & Time & Para & Mem  & Time          \\ \hline
CTC      & none & 0.1 &  3.1 & none& 47.8&   16.2       \\  
Attention& 2.8 & 6.6  & 78.9 &17.2 & 143.6 & 85.5  \\
ACE      & {\bf none} & {\bf 0.02} & {\bf $<$0.1} & {\bf none} & {\bf 4.2}& {\bf $<$0.1}        \\ 
\hline                                                                              
\end{tabular}
\end{table}

\section{Conclusion}
In this paper, a novel and straightforward ACE loss function is proposed for sequence recognition problem with competitive performance to CTC and attention. 
Owing to its simplicity, the ACE loss function is easy to employ by simply replacing CTC with ACE, quick to implement with only four basic formulas, fast to infer and back-propagate at approximately $O(1)$ in parallel, and exhibits marginal memory requirement.
Two following effective properties of ACE loss function are also investigated: (1) it can easily handle 2D prediction problem with marginal adaption and (2) it does not require character-order information for supervision, which allows it to advance beyond sequence recognition problem, e.g., counting problem.


\section*{Acknowledgments}
\small
This research is supported in part by GD-NSF (no. 2017A030312006), the National Key Research and Development Program  of China (No. 2016YFB1001405), NSFC (Grant No.: 61673182, 61771199), and GDSTP (Grant No.:2017A010101027) , GZSTP(no. 201704020134).

\end{document}